\documentclass[conference,letterpaper]{IEEEtran}
\usepackage{cite}
\usepackage{amsmath,amssymb,amsfonts}
\usepackage{graphicx}
\usepackage{booktabs}
\usepackage{textcomp}
\usepackage{dblfloatfix}
\usepackage{array}
\usepackage{float}
\usepackage{adjustbox}
\usepackage{placeins}
\usepackage{xcolor}
\usepackage{balance}
\begin{document}

\title{A Graph Neural Network--Guided Genetic Algorithm for Physical Internet Supply Chain Optimization under Cost Uncertainty}

\author{
\IEEEauthorblockN{Faezeh Ardali}
\IEEEauthorblockA{Department of Mechanical and Industrial Engineering\\
Louisiana State University, Baton Rouge, LA, USA\\
fardal1@lsu.edu}
\and
\IEEEauthorblockN{Gerald M. Knapp}
\IEEEauthorblockA{Department of Mechanical and Industrial Engineering\\
Louisiana State University, Baton Rouge, LA, USA\\
gknapp@lsu.edu}
}

\maketitle

\begin{abstract}
Inventory and distribution planning in Physical Internet networks requires coordinating factory–hub assignments, factory supply, lateral transshipment among collaborative hubs, retailer deliveries, and shortages. The problem combines discrete assignment decisions with interdependent continuous flows, while uncertain operating costs make robust planning more difficult. This study formulates deterministic and min–max regret models for a three-echelon network of factories, hubs, and retailers and develops a graph neural network-guided genetic algorithm (GNN–GA) for the assignment decisions. The GNN estimates hub-specific factory-selection probabilities that are used to construct the initial GA population and adapt mutation according to prediction uncertainty. Each previously unseen candidate assignment is evaluated by solving the remaining continuous-flow problem to LP optimality. Simulated annealing, a standard GA, and GNN–GA are compared on 15 instances using matched random seeds and fixed limits on distinct assignment evaluations. Because the evaluation budgets for test Instances 13–15 are smaller than the nominal population size, these experiments primarily assess the quality of learned initialization rather than multi-generation evolutionary search. A separate 400-evaluation experiment on exact test Instance 13 permits three complete offspring generations and a partial fourth pass, with GNN–GA outperforming GA in all 10 matched runs. Three independently generated exact-solvable instances provide a separate test of transfer. Ablation results show that learned initialization provides most of the improvement, while entropy-guided mutation has a smaller, instance-dependent effect. Per-instance solution times include GNN inference and search but exclude model training and one-time model setup.
\end{abstract}


\section*{Nomenclature}
\footnotesize
\setlength{\tabcolsep}{2pt}
\begin{tabular}{@{}>{\raggedright\arraybackslash}p{0.27\columnwidth}>{\raggedright\arraybackslash}p{0.66\columnwidth}@{}}
$O,H,R,\mathcal S$ & Factories, hubs, retailers, and cost scenarios; indices $o,h,h',r,s$.\\
$O_h$ & Nonempty set of factories eligible to supply hub $h$.\\
$\mathcal A^H$ & Valid directed hub-to-hub arcs $(h,h')$, $h\ne h'$.\\
$CF,FC,CH$ & Factory-to-hub supply, fixed assignment, and hub-transshipment costs.\\
$CR,CS$ & Retailer-delivery and shortage costs; superscript $s$ denotes scenario values.\\
$Dem_r,Cap_o$ & Retailer demand and factory capacity.\\
$I_h^0,WHC_h$ & Initial hub inventory and factory-inbound capacity.\\
$X,QF,QH$ & Factory assignment, factory flow, and hub-transshipment variables.\\
$QR,U,\mathcal R$ & Retailer flow, shortage, and maximum regret.\\
\end{tabular}
\normalsize

\section{Introduction}
The Physical Internet (PI) envisions logistics as an interconnected, standardized, and resource-sharing network that supports coordinated inventory and distribution planning \cite{montreuil2011,pan2015,treiblmaier2020}. Such coordination remains difficult because factory assignment, capacity allocation, lateral hub transshipment, retailer service, and shortage decisions are interdependent. These decisions become more challenging when production, transportation, assignment, and shortage costs vary. When probabilities for future cost realizations are unavailable,
min--max regret optimization limits the largest deviation from the
scenario-specific optimal outcome
\cite{bertsimas2004,bental2009,aissi2009}.

Integrated production--distribution and multi-echelon models demonstrate the value of coordinating assignment, allocation, and flow decisions \cite{azani2025shrimp,darvish2021}.Related manufacturing studies have
also used rolling-horizon optimization to coordinate operational decisions
under uncertain or delayed information \cite{parmar2026}. Within the PI literature, previous studies have examined digital interoperability, resilience planning, and hybrid intelligent supply-chain management \cite{pan2021digital,peng2021resilience,yan2023enhancing}. These studies establish the value of interconnected planning, but they provide limited support for solving a model that simultaneously includes factory assignment, lateral transshipment, retailer delivery, shortage, and coordinated cost scenarios.

Learning-based methods offer a way to guide difficult combinatorial decisions without replacing the optimization model that enforces feasibility \cite{bengio2021,karimi2022}. GNNs are particularly suitable for relational decisions in networked systems \cite{kipf2017,wu2021gnn,dai2017learning}. Related learning-based approaches have been applied to scheduling and routing
\cite{ardali2026drlt,kool2019attention}, as well as resource allocation and
post-disaster decision making
\cite{soleymani2024,torkaman2026hurricane,amani2026earthquake}.
Other studies have explored learning-based optimization for restoration and
broader combinatorial decision problems
\cite{amani2026storm,mazyavkina2021survey,rajabi2026dqaoa}.
In the proposed method, however, the GNN does not predict shipment quantities
or replace mathematical optimization. It only guides the discrete
factory--hub assignment search; all continuous flows are still determined by
an exact linear program for each distinct assignment.. In the proposed method, however, the GNN does not predict shipment quantities or replace mathematical optimization. It only guides the discrete factory--hub assignment search; all continuous flows are still determined by an exact linear program for each distinct assignment.

This study makes four contributions. First, it formulates deterministic and min--max regret PI models that integrate single-source factory assignment, factory supply, lateral hub transshipment, retailer delivery, and shortage. Second, it develops a GNN-guided GA in which learned assignment probabilities improve population initialization and adapt mutation to prediction uncertainty. Third, it evaluates the method against simulated annealing (SA), a standard GA, and a non-learning cost-ranked warm start under matched seeds and common evaluation limits. Fourth, it distinguishes the instances used for GNN development from held-out and independently generated tests and supplements the tight-budget held-out study with a larger-budget, multi-generation experiment.

\section{Physical Internet Optimization Model and Hybrid Solution Method}
\subsection{Deterministic and Min--Max Regret Models}
A single product moves from factories to hubs and then to retailers. Each hub is assigned to exactly one eligible factory, unmet demand is penalized, and unused inventory may remain at the end of the planning period. Under nominal costs, the model is
{\small
\begin{align}
\min Z ={}&
\sum_{o\in O}\sum_{h\in H}
\left(FC_{oh}X_{oh}+CF_{oh}QF_{oh}\right)
\notag\\
&+\sum_{(h,h')\in\mathcal A^H}
CH_{hh'}QH_{hh'}
+\sum_{h\in H}\sum_{r\in R}
CR_{hr}QR_{hr}
\notag\\
&+\sum_{r\in R}CS_rU_r
\label{eq:obj}\\
\text{s.t.}\quad
&\sum_{h\in H}QF_{oh}\leq Cap_o,
\qquad \forall o,
\label{eq:factory-capacity}\\
&\sum_{o\in O_h}X_{oh}=1,
\qquad \forall h,
\label{eq:single-sourcing}\\
&X_{oh}=0,\quad QF_{oh}=0,
\qquad \forall h\in H,\;o\in O\setminus O_h,
\label{eq:eligibility}\\
&QF_{oh}\leq WHC_hX_{oh},
\qquad \forall o,h,
\label{eq:shipment-linking}\\
&\sum_{o\in O}QF_{oh}
+\sum_{\substack{h'\in H:\\(h',h)\in\mathcal A^H}}QH_{h'h}
+I_h^0
\notag\\[-1mm]
&\qquad\geq
\sum_{r\in R}QR_{hr}
+\sum_{\substack{h'\in H:\\(h,h')\in\mathcal A^H}}QH_{hh'},
\qquad \forall h,
\label{eq:hub-balance}\\
&\sum_{h\in H}QR_{hr}+U_r=Dem_r,
\qquad \forall r,
\label{eq:demand-balance}\\
&X_{oh}\in\{0,1\},
\qquad \forall o,h,
\label{eq:binary-domain}\\
&QF_{oh},\,QR_{hr},\,U_r\geq0,
\qquad \forall o,h,r,
\notag\\[-1mm]
&QH_{hh'}\geq0,
\qquad \forall (h,h')\in\mathcal A^H.
\label{eq:continuous-domain}
\end{align}
}
The objective combines fixed factory assignments, factory supply, lateral transshipment, retailer delivery, and unmet-demand penalties. The constraints enforce factory capacity, single sourcing, factory eligibility, assignment-dependent inbound flow, hub material balance, and retailer demand balance. Because the hub balance is an inequality, unused terminal inventory is permitted. The resulting formulation is a mixed-integer linear program.

Cost uncertainty is represented by coordinated lower, nominal, and upper scenarios. Supply and fixed-assignment costs use multipliers $0.8/1/1.2$, transshipment $0.5/1/1.5$, delivery $0.9/1/1.1$, and shortage $0.7/1/1.3$. These scenarios are a coordinated stress test rather than a complete uncertainty set containing every mixed realization. For feasible decision vector $x$, scenario cost $Z_s(x)$, and scenario optimum $Z_s^*$, the robust plan solves
\begin{equation}
\min_{x\in\mathcal X,\mathcal R}\mathcal R
\quad\mathrm{s.t.}\quad
\mathcal R\ge Z_s(x)-Z_s^*,\ \forall s.
\label{eq:regret}
\end{equation}
All assignments and flows are selected before the scenario is known, and $\mathcal R$ is the largest regret over the three scenarios.

\subsection{Graph Neural Network--Guided Genetic Algorithm}
Factory--hub assignments define the discrete search space, while exact
optimization determines the continuous flows. The heterogeneous graph
contains factory, hub, and retailer nodes. Factory--hub and hub--retailer
edges are complete in the generated benchmarks, and directed hub--hub
edges include every ordered pair except self-arcs. Type-specific encoders
map factory, hub, and retailer features to a common hidden space.
Relation-specific multilayer perceptrons form messages for factory--hub,
hub--factory, hub--retailer, retailer--hub, and hub--hub relations.
Messages are mean-aggregated at each receiving node and combined through
residual, layer-normalized updates with ReLU activation for two rounds with hidden dimension
32 and zero dropout. Features describe capacity, inventory, demand,
shortage penalties, eligibility, and local cost information. Continuous
features are standardized using statistics fitted only on training
Instances 1--9.

The final factory--hub edge scorer produces a compatibility logit $s_{oh}$
for each eligible pair. The corresponding assignment probability is
\begin{equation}
p_{oh}=\frac{\exp(s_{oh})}
{\sum_{o'\in O_h}\exp(s_{o'h})}.
\label{eq:prob}
\end{equation}
Because the exact deterministic and regret assignments coincide for
Instances 1--12, one shared assignment model is used for both formulations.
Optimal factory assignments obtained from the complete models for Instances 1--9 provide the categorical training labels. Instances 10–12 were excluded from gradient updates but were used for validation, early stopping, and selection of the network architecture and guided-population fraction. The GNN was implemented in PyTorch and trained with AdamW using a learning rate of 0.01, weight decay of $10^{-4}$, cross-entropy loss, gradient clipping at 5, a 100-epoch limit, early-stopping patience of 25, and random seed 3101. The selected model achieved 75.56\% hub-level accuracy on the validation instances; five independent training runs produced $73.78\%\pm3.76\%$.

A chromosome $\mathbf c=(o_h)_{h\in H}$ contains one eligible factory index for each hub. Once $\mathbf c$ is fixed, all binary assignments are known and the remaining deterministic or regret problem is a continuous linear program. Each previously unseen chromosome is evaluated with the HiGHS solver, and its objective is stored so that repeated chromosomes do not require another solve. The linear-program structure is kept in memory and only the assignment-dependent bounds and right-hand sides are updated. Thus, the GNN ranks discrete choices but does not determine shipment quantities, relax constraints, or accept a candidate without exact continuous-flow evaluation.

Standard GA samples each gene uniformly over $O_h$. GNN--GA includes the most likely chromosome, samples 80\% of the effective population from \eqref{eq:prob}, and fills the remainder uniformly. With distinct-evaluation budget $B$, nominal population $P_{\rm nom}$, and assignment-space size $|\Omega|$, the implemented population is
\begin{equation}
P_{\rm eff}=\min\{P_{\rm nom},B,|\Omega|\}.
\label{eq:peff}
\end{equation}
Thus, when $B<P_{\rm nom}$, the initial population is explicitly truncated rather than evaluated beyond the budget. Prediction entropy and mutation are defined explicitly for both nonsingleton and singleton eligibility:
\begin{align}
\widetilde H_h&=
\begin{cases}
-\dfrac{\sum_{o\in O_h}p_{oh}\log p_{oh}}{\log|O_h|},& |O_h|>1,\\
0,& |O_h|=1,
\end{cases}\notag\\[-1mm]
\mu_h&=
\begin{cases}
\min\{0.20,0.05(0.5+1.5\widetilde H_h)\},& |O_h|>1,\\
0,& |O_h|=1.
\end{cases}
\label{eq:mutation}
\end{align}
Tournament selection of size 3, uniform crossover with probability 0.8, and 5\% elitism are identical in GA and GNN--GA. Standard GA uses mutation probability 0.05 only for genes satisfying $|O_h|>1$. SA selects a hub from $H_{\rm mov}=\{h\in H:|O_h|>1\}$ and changes its supplier. If $H_{\rm mov}=\varnothing$, SA evaluates the unique chromosome once and terminates. If $|\Omega|=1$, GA and GNN--GA likewise evaluate the
unique chromosome once and terminate without assignment-changing
crossover or mutation.

\begin{table}[!t]
\centering
\caption{Common Search and Implementation Settings}
\label{tab:settings}
\scriptsize\setlength{\tabcolsep}{3pt}
\begin{tabular}{@{}ll@{}}
\toprule
Setting & Value\\\midrule
Seeds & 1001--1020; sensitivity: 1001--1010\\
Nominal population & 50 (Inst. 3--9), 100 (Inst. 10--15)\\
Effective population & $\min(P_{\rm nom},B,|\Omega|)$\\
Selection/crossover & Tournament 3 / uniform 0.8\\
Elitism / fixed mutation & 5\% / 0.05 per movable gene\\
GNN initialization & 80\% guided (greedy included), 20\% uniform \\
Stopping criterion & Maximum distinct LP evaluations\\
\bottomrule
\end{tabular}
\end{table}

\section{Experimental Setup and Evaluation Protocol}
Fifteen benchmark instances increase from $(|O|,|H|,|R|)=(1,3,2)$ to $(15,90,700)$. Demand is sampled from 80--120; supply costs from 80--140; fixed assignment costs from 8000--20000; transshipment costs from 10--30; delivery costs from 20--60; and shortage penalties from 250--400. Factory capacity and hub inbound capacity are proportional allocations of 120\% and 110\% of total demand. A fixed pseudorandom seed of 42 was used to generate the benchmark set. All factory--hub pairs are eligible, giving an eligibility density of 1.00. The hub--hub network contains every directed arc except self-arcs, and every hub can serve every retailer.

Instances 1--9 provide GNN training labels, Instances 10--12 are used for model selection, and Instances 13--15 are excluded from both stages. Exact references are available through Instance 13; Instances 14--15 use best-known incumbents and are therefore not reported with exact optimality gaps. Instance 1 has dimensions $1/3/2$ with $O_1=O_2=O_3=\{1\}$ and unique chromosome $(1,1,1)$. Instance 2 has dimensions $1/5/7$ with $O_1=\cdots=O_5=\{1\}$ and unique chromosome $(1,1,1,1,1)$. They are used as fixed-assignment verification cases rather than stochastic-search comparisons.

The tight-budget distinct-evaluation budgets are 64 and 256 for Instances 3--4; 500 for 5--10; and 250, 150, 75, 25, and 10 for 11--15. Consequently, the held-out populations are truncated from the nominal 100 to 75, 25, and 10. The entire budget is consumed by initialization, so these runs complete no offspring generation. Their role is therefore to test the quality of learned initialization under severe limits, not the complete evolutionary process.

\begin{table}[!t]
\centering
\caption{Data Split and Held-Out Budget Interpretation}
\label{tab:protocol}
\scriptsize\setlength{\tabcolsep}{2.6pt}
\begin{tabular}{@{}clrrr@{}}
\toprule
Inst. & Role & $B$ & $P_{\rm eff}$ & Full gen.\\\midrule
1--2 & Training/fixed assignment & 1 & 1 & 0\\
3--9 & Training & 64--500 & 50 & search dependent\\
10--12 & Model selection & 500/250/150 & 100 & search dependent\\
13 & Held out, tight budget & 75 & 75 & 0\\
14 & Held out, tight budget & 25 & 25 & 0\\
15 & Held out, tight budget & 10 & 10 & 0\\
13 & Sensitivity & 400 & 100 & 3 + partial\\
\bottomrule
\end{tabular}
\end{table}

To evaluate multi-generation behavior, a separate experiment increases the evaluation limit for exact held-out Instance 13 to 400, uses a population of 100, and applies 10 matched seeds. This permits three complete offspring generations and a fourth, budget-limited generation. Three additional exact-solvable instances with dimensions $(3,10,25)$, $(4,12,30)$, and $(5,15,40)$ were generated with independent seeds 202601--202603 after the model and search settings had been selected. No normalization statistic, model parameter, or search setting was changed after observing these tests.

The optimization models, data-generation procedure, and search algorithms were implemented in Python. PyTorch was used for GNN training and inference, and the HiGHS solver accessed through SciPy was used for the complete mixed-integer models and fixed-assignment linear programs. All solver calls used one thread. The reported per-instance solution time includes population operations, cache lookups, LP solves, and GNN inference when applicable. It excludes GNN training, loading the trained weights, and constructing the solver model once for each instance; the time ratios therefore compare the repeated solution stage under the selected evaluation limits rather than complete end-to-end development time. Exact nonzero references use percentage gap, while zero or near-zero regret references use absolute deviation. For each nonsingleton instance--formulation pair, matched-seed
Wilcoxon tests compare the three method pairs, and Holm correction
is applied within that three-test family. The 26 three-method Friedman
tests are Holm-corrected across the 26 nonsingleton
instance--formulation combinations. Exact references are obtained from the complete deterministic or regret mixed-integer model rather than from the best heuristic run. Instances 3--4 are additionally verified by enumerating all 64 and 256 assignment chromosomes. For sampled assignments, the fixed-chromosome LP is cross-checked against the complete model with its assignment variables fixed, and regret is independently reconstructed as $\max_s\{Z_s(\mathbf c)-Z_s^*\}$. All verification differences are below $10^{-6}$. The instance generator, implementation, trained model parameters,
and raw run-level experimental results are available from the authors
upon reasonable request.

\section{Results and Discussion}
\subsection{Model-Development and Tight-Budget Held-Out Results}
Table~\ref{tab:development} summarizes the model-development instances. GNN--GA reached the exact deterministic reference in every run on training Instances 3--9. It also produced the lowest mean objective on validation Instances 10--12, although those instances influenced model selection and are not independent test evidence. Regret results are reported as objective values because several exact references are zero or close to zero.

\begin{table*}[!t]
\centering
\caption{Model-Development Results for Instances 3--12}
\label{tab:development}
\scriptsize\setlength{\tabcolsep}{3.0pt}
\begin{adjustbox}{max width=\textwidth}
\begin{tabular}{crrrcrrrr}
\toprule
& \multicolumn{3}{c}{Deterministic mean gap (\%)} && \multicolumn{4}{c}{Min--max regret mean objective}\\
\cmidrule(lr){2-4}\cmidrule(lr){6-9}
Inst. & SA & GA & GNN--GA && Ref. & SA & GA & GNN--GA\\\midrule
3 & 0.00 & 0.00 & \textbf{0.00} && 85.94 & 85.94 & 85.94 & \textbf{85.94}\\
4 & 0.00 & 0.00 & \textbf{0.00} && 208.57 & 208.57 & 208.57 & \textbf{208.57}\\
5 & 0.00 & 0.00 & \textbf{0.00} && 0.00 & 0.00 & 0.00 & \textbf{0.00}\\
6 & 0.13 & 0.00 & \textbf{0.00} && 41.86 & 41.86 & 41.86 & \textbf{41.86}\\
7 & 1.41 & 0.20 & \textbf{0.00} && 30.56 & 74.87 & 1584.71 & \textbf{30.56}\\
8 & 3.57 & 0.56 & \textbf{0.00} && 0.35 & 0.35 & 4023.54 & \textbf{0.35}\\
9 & 4.91 & 1.61 & \textbf{0.00} && 6.50 & 4942.61 & 14330.62 & \textbf{6.50}\\
10 & 7.21 & 5.32 & \textbf{0.18} && 3.15 & 14651.50 & 48890.89 & \textbf{1714.66}\\
11 & 12.79 & 10.73 & \textbf{2.02} && 0.05 & 119145.27 & 152123.05 & \textbf{28276.16}\\
12 & 16.55 & 13.88 & \textbf{1.75} && 0.00 & 280265.08 & 278669.13 & \textbf{34846.69}\\
\bottomrule
\end{tabular}
\end{adjustbox}
\vspace{0.3mm}\parbox{0.96\textwidth}{\footnotesize Instances 3--9 are training benchmarks and 10--12 influence model selection; neither block is independent test evidence.}
\end{table*}

Table~\ref{tab:heldout} compares the three search methods and a non-learning warm start on the tight-budget held-out budgets. The cost-ranked baseline uses only factory--hub supply and fixed-assignment ranks. GNN--GA wins all 20 paired runs against this baseline on Instances 13 and 15, whereas the cost warm start wins all 20 on Instance 14. This exception is important: the learned model is usually beneficial, but it is not uniformly better than a simple cost rule. Because no offspring generation occurs, full GNN--GA and the initialization-only variant are identical in these runs.

\begin{table*}[!t]
\centering
\caption{Held-Out Tight-Budget Results and Cost-Ranked Warm Start (Mean $\pm$ SD)}
\label{tab:heldout}
\scriptsize\setlength{\tabcolsep}{2.7pt}
\begin{adjustbox}{max width=\textwidth}
\begin{tabular}{ccrrrrrcc}
\toprule
Inst. & Form. & Baseline & SA & GA & Full GNN--GA & Cost warm start & Full/Cost wins & $p_{\rm Holm}$\\\midrule
13 & Det. & 2,531,549 & 3,036,226 $\pm$ 45,741 & 2,982,500 $\pm$ 25,065 & 2,561,756 $\pm$ 3,871 & 2,572,443 $\pm$ 0 & 20/0 & 8.8e-05 \\
13 & Reg. & 0 & 556,259 $\pm$ 46,720 & 541,135 $\pm$ 30,094 & 36,213 $\pm$ 4,650 & 49,005 $\pm$ 0 & 20/0 & 1.1e-05 \\
14 & Det. & 3,785,785 & 4,608,176 $\pm$ 62,110 & 4,508,967 $\pm$ 38,477 & 3,861,512 $\pm$ 11,555 & 3,828,878 $\pm$ 0 & 0/20 & 1.1e-05 \\
14 & Reg. & 0 & 982,293 $\pm$ 74,547 & 867,983 $\pm$ 46,163 & 91,123 $\pm$ 14,131 & 51,854 $\pm$ 0 & 0/20 & 1.1e-05 \\
15 & Det. & 8,000,214 & 9,877,367 $\pm$ 157,119 & 9,646,684 $\pm$ 77,708 & 8,359,145 $\pm$ 49,668 & 8,717,492 $\pm$ 0 & 20/0 & 1.1e-05 \\
15 & Reg. & 0 & 2,239,419 $\pm$ 189,679 & 1,976,882 $\pm$ 94,356 & 437,636 $\pm$ 68,336 & 960,530 $\pm$ 0 & 20/0 & 1.1e-05 \\
\bottomrule
\end{tabular}
\end{adjustbox}
\vspace{0.3mm}

\parbox{0.96\textwidth}{\footnotesize
The Instance~13 baseline is exact. Instance~14--15 deterministic and
scenario baselines are best-known; their reported regret objectives are
incumbent-relative surrogate regrets rather than verified exact
min--max regrets.
}
\end{table*}

\subsection{Multi-Generation Sensitivity and Independent Exact Tests}
The 400-evaluation Instance-13 experiment evaluates whether the learned advantage remains after the algorithms proceed beyond initialization. As shown in Table~\ref{tab:sensitivity}, GNN--GA outperforms GA in every matched run for both formulations after three complete offspring generations and a partial fourth pass. The result supports the full evolutionary method under this larger budget, but it does not imply that additional generations will always increase the performance difference.

\begin{table*}[!t]
\centering
\caption{Exact Held-Out Instance 13: Sensitivity with $B=400$}
\label{tab:sensitivity}
\scriptsize\setlength{\tabcolsep}{3pt}
\begin{adjustbox}{max width=\textwidth}
\begin{tabular}{ccrrrrccc}
\toprule
Inst. & Form. & GA mean $\pm$ SD & GNN--GA mean $\pm$ SD & GA gen. & GNN gen. & GNN wins & $p_{\rm Holm}$ & Ref. status\\\midrule
13 & Det. & 2,903,866 $\pm$ 20,732 & 2,558,140 $\pm$ 3,030 & 3 + part. & 3 + part. & 10/10 & 0.0039 & Exact \\
13 & Reg. & 437,447 $\pm$ 22,669 & 30,897 $\pm$ 3,389 & 3 + part. & 3 + part. & 10/10 & 0.0039 & Exact \\
\bottomrule
\end{tabular}
\end{adjustbox}
\end{table*}

The three independent exact tests provide a separate check of transfer after model selection is complete. All methods solve T1 exactly. GNN--GA has the lowest mean on both T2 formulations and deterministic T3, whereas SA is strongest on T3 regret and reaches the exact reference in 10 of 20 runs. These mixed outcomes prevent a claim of universal dominance and indicate that difficult regret landscapes may require a larger and more formulation-diverse training corpus.

\begin{table*}[!t]
\centering
\caption{Independent Exact Tests
(Mean $\pm$ SD; Exact-Success Counts by Method)}
\vspace{-5pt}
\label{tab:exactnew}
\scriptsize\setlength{\tabcolsep}{2.7pt}
\begin{adjustbox}{max width=\textwidth}
\begin{tabular}{ccccrrrr}
\toprule
Test & $O/H/R$ & Form. & Ref. & SA & GA & GNN--GA & Exact successes (SA/GA/GNN)\\\midrule
T1 & 3/10/25 & Det. & 405,205 & 405,205 $\pm$ 0 & 405,205 $\pm$ 0 & 405,205 $\pm$ 0 & 20/20/20\\
T1 & 3/10/25 & Reg. & 553.62 & 553.62 $\pm$ 0 & 553.62 $\pm$ 0 & 553.62 $\pm$ 0 & 20/20/20\\
T2 & 4/12/30 & Det. & 468,856 & 472,987 $\pm$ 2,896 & 469,726 $\pm$ 1,225 & \textbf{469,188 $\pm$ 193} & 0/3/2\\
T2 & 4/12/30 & Reg. & 78.24 & 802.62 $\pm$ 1,411 & 763.75 $\pm$ 1,038 & \textbf{430.11 $\pm$ 174} & 5/5/0\\
T3 & 5/15/40 & Det. & 618,449 & 646,369 $\pm$ 11,140 & 627,125 $\pm$ 5,000 & \textbf{626,161 $\pm$ 3,707} & 0/1/0\\
T3 & 5/15/40 & Reg. & 31.67 & \textbf{2,230 $\pm$ 2,963} & 13,925 $\pm$ 7,167 & 9,351 $\pm$ 4,189 & 10/0/0\\
\bottomrule
\end{tabular}
\end{adjustbox}
\end{table*}

\begin{figure*}[!t]
\centering
\includegraphics[width=0.86\textwidth]{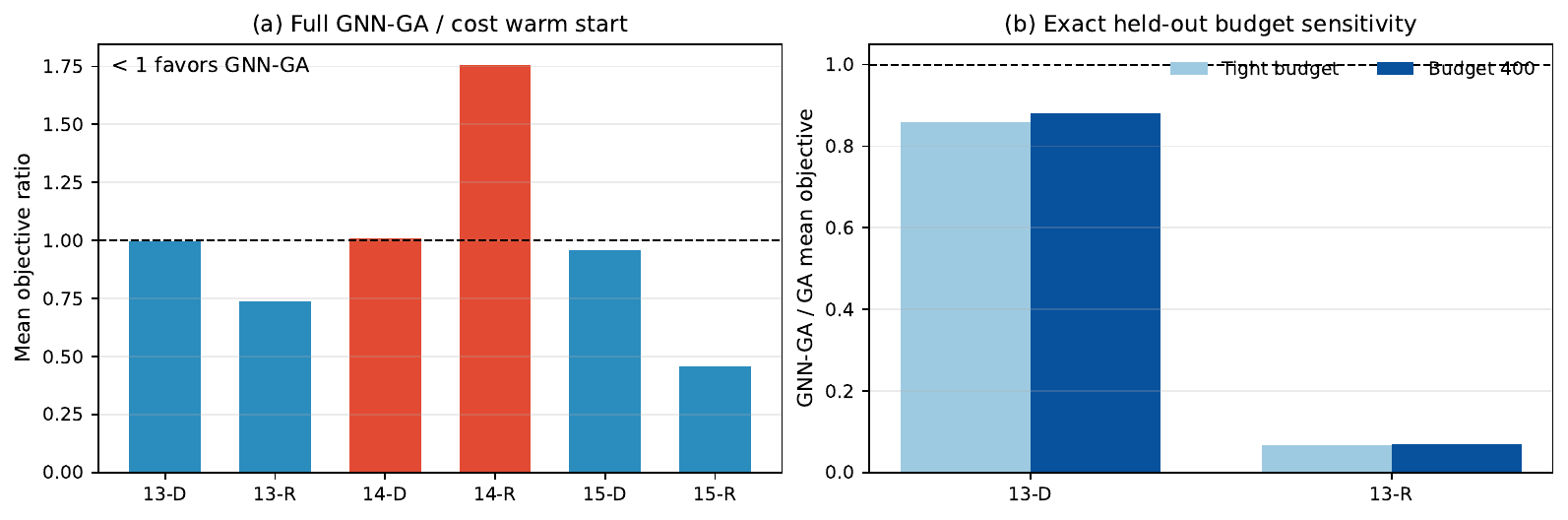}
\caption{Held-out comparisons. Left: full GNN--GA relative to the non-learning cost warm start under the tight budgets. Right: GNN--GA relative to GA under the tight-budget and 400-evaluation Instance-13 budgets; values below one favor GNN--GA.}
\label{fig:heldout}
\end{figure*}

\subsection{Ablation, Runtime, and Statistical Interpretation}
Ablation across the 26 nonsingleton tight budget comparisons shows that full GNN--GA is better/tied/worse than the no-GNN control in 20/6/0 cases, initialization-only in 4/20/2, and the cost warm start in 16/8/2. Learned initialization is therefore the main source of improvement. Entropy-dependent mutation offers a smaller contribution and requires sufficient budget for offspring generations; it cannot be assessed from the tight budget held-out runs, where the initialization consumes the full budget.

\begin{table}[!t]
\centering
\caption{Ablation Outcomes Across 26 Nonsingleton Comparisons}
\label{tab:ablation}
\scriptsize\setlength{\tabcolsep}{3pt}
\begin{tabular}{@{}lccc@{}}
\toprule
Full GNN--GA versus & Better & Tie & Worse\\\midrule
No-GNN control & 20 & 6 & 0\\
Initialization only & 4 & 20 & 2\\
Cost warm start & 16 & 8 & 2\\
\bottomrule
\end{tabular}
\end{table}

With all three methods evaluated using the same single-threaded solver implementation, the held-out per-instance time ratios are $T_{SA}/T_{GNN}=3.476$ and $T_{GA}/T_{GNN}=5.265$ under the selected evaluation limits. These ratios exclude GNN training, loading the trained weights, and one-time model construction, and therefore describe only the repeated solution stage. On small instances, the inference cost is not fully offset and GNN--GA can be slower than SA.

Paired Wilcoxon comparisons use matched seeds and the correction
families defined in Section III. The tight budget held-out comparisons with SA and GA are significant, but they describe tight-budget performance dominated by initialization. The expanded-budget Instance-13 results evaluate the algorithms after several generations, while T1--T3 provide independent exact tests. Repeating the SA and GA experiments with the common solver implementation reproduced 1198 of 1200 earlier objectives within $10^{-6}$. The two differences occurred in GA regret runs for Instance~8 and
resulted from alternative numerical tie resolution; they did not
change the aggregate conclusions.

Across held-out Instances 13--15, all 12 paired comparisons of GNN--GA with SA or GA are significant after Holm correction. Across the 26 nonsingleton instance--formulation combinations, the three-method Friedman test is significant in 19 cases. These statistics support the observed differences under the tested evaluation limits. Their interpretation follows the experimental split: Instances 3--9 are training results, 10--12 are model-selection results, Instance 13 is an exact held-out comparison, Instances 14--15 are descriptive comparisons against best-known incumbents, and T1--T3 are independent exact tests. Exact-gap claims are restricted to cases with verified references.

\section{Conclusion}
This study developed deterministic and min--max regret models for coordinated factory assignment, supply, lateral transshipment, retailer delivery, and shortage decisions in a Physical Internet network. The proposed GNN--GA learns factory--hub assignment probabilities, uses them to construct promising initial populations, and adjusts mutation according to prediction uncertainty. Exact continuous-flow optimization evaluates every distinct chromosome and preserves feasibility and interpretability.

The experimental results show that learned initialization is the dominant benefit. Under the tight-budget held-out experiments, the population is truncated and the results should be interpreted as an initialization study. When exact held-out Instance 13 is given enough budget for several generations, GNN--GA still outperforms GA in every paired run. The independent exact tests provide mixed but informative transfer results: GNN--GA performs best on several difficult cases, while SA is strongest on one regret test. The method is therefore frequently useful but not uniformly dominant.

The conclusions remain limited to synthetic single-period networks, complete eligibility and connectivity, and three coordinated cost scenarios. Future work should expand the exact held-out collection, introduce sparse and externally derived topologies, train on more formulation-diverse assignment labels, and study dynamic or mixed uncertainty.


\end{document}